\documentclass[graybox]{svmult}

\usepackage{type1cm}        % activate if the above 3 fonts are
\usepackage{makeidx}         % allows index generation
\usepackage{graphicx}        % standard LaTeX graphics tool
\usepackage{multicol}        % used for the two-column index
\usepackage[bottom]{footmisc}% places footnotes at page bottom
\usepackage{booktabs}

\usepackage{newtxtext}       % 
\usepackage{newtxmath}       % selects Times Roman as basic font

\def\R{\mathbb{R}}
\def\N{\mathbb{N}}
\def\dom{\mathop{\rm dom}}
\def\q{\boldsymbol{q}}
\def\weight{\varpi}
\def\friction{\eta}
\makeindex             % used for the subject index
\begin{document}

\title*{Least Action Principles and \\ Well-Posed Learning Problems}
\author{Alessandro Betti and Marco Gori}
\institute{Alessandro Betti \at DINFO, Via di S. Marta, 3 Firenze, Italia,
\email{alessandro.betti@unifi.it}
\and Marco Gori \at DIISM,  via Roma, 56, Siena, Italia
\email{marco@dii.unisi.it}}
\maketitle

\abstract*{Each chapter should be preceded by an abstract (no more
than 200 words) that summarizes the content. The abstract will appear
\textit{online} at \url{www.SpringerLink.com} and be available with
unrestricted access. This allows unregistered users to read the
abstract as a teaser for the complete chapter.  Please use the
'starred' version of the \texttt{abstract} command for typesetting the
text of the online abstracts (cf. source file of this chapter template
\texttt{abstract}) and include them with the source files of your
manuscript. Use the plain \texttt{abstract} command if the abstract is
also to appear in the printed version of the book.}

\abstract{Machine Learning algorithms are typically regarded as
appropriate optimization schemes for minimizing risk functions 
that are constructed on the training set, which conveys statistical
flavor to the corresponding learning problem. When the focus is 
shifted on perception, which is inherently interwound with time, recent
alternative formulations of learning have been proposed that 
rely on the principle of Least Cognitive Action, 
which very much reminds us of the Least Action Principle in mechanics. 
In this paper, we discuss different forms of the cognitive action and
show the well-posedness of learning. In particular, unlike the special
case of the action in mechanics, where the stationarity is typically gained
on saddle points, we prove the existence  of the minimum of a special form 
of cognitive action, which yields forth-order differential equations of learning.
We also briefly discuss the dissipative behavior of these equations that 
turns out to characterize the process of learning.}

\section{Introduction}
Whenever a learning process is embedded in a temporal environment; i.e. the
data presented to the agent has a temporal structure (video and audio signals
for example) it seems natural to define the learning process directly
through the definition of a suitable 
temporal dynamics. In other words one might start to think that the updating
of the model's parameters, which is what we usually call ``learning'', must
be synced with the temporal structure of data. This suggests 
investigating the continuous map $t\mapsto w(t)$
as a response to the input  $u(t)$, thus regarding $t$ as time and not simply
an iteration index of popular machine learning algorithms.

In order to be able to select the correct dynamics of the weights of an agent
we believe that a functional formulation of the problem is
particularly useful. 
For example, the Lagrangian formulation of physical theories offers
the possibility of imposing all the symmetries of a theory simply adding to
the Lagrangian terms that satisfy such symmetry (see for example~\cite{weinberg}).
In the same way \cite{tnnls-vision}, this approach makes it easier to
incorporate constraints on the dynamic of the learned weights.
A  variational approach based on an integral functional like the
action of classical mechanics can be conceived which specifies 
in one single scalar function (what
in mechanics is called the Lagrangian) both the ``static'' goodness criterion,
the potential, and the dynamical part of learning by a 
kinetic term~\cite{tcs}.

For example, consider a classical batch problem in machine learning where
the functional risk has been approximated with a function $V(w)$.
As we will discuss in Section~\ref{sec:2} we can find appropriate functional
indexes that have as stationarity condition the following differential equation
\begin{equation} m \ddot w +\friction\dot w +\nabla V(w)=0
\qquad m, \friction>0.
\label{heavy}\end{equation}
This equation can be considered as the continuous form of a classic
multistep first order method (see~\cite{polyak}) known as the {\it heavy ball} method.
The name of this method derive from the fact that
Eq.~\eqref{heavy} can be interpreted as the equation of motion of an
heavy ball with friction subject to the potential $V(w)$. Equation~\eqref{heavy} is
also closely related to the continuous approximation of other first order methods 
(see~\cite{boyd}).
More directly
in the case $m\to 0$ and $\friction$ fixed we get the continuous version of
a plain gradient descent method with learning rate $1/\friction$:
\[\dot w=-\frac1 \friction\nabla V(w). \]
Notice the importance of the first order term in Eq.~\eqref{heavy}; without
dissipation we wouldn't be able to recover the classical gradient descent
method. Even worse, in general without the presence of the $\friction$ term
there is no hope for the dynamic to reach a stationary point of $V$. Indeed,
broadly speaking, since in that case the mechanical energy would be conserved
lower values of $V$ correspond to higher values of the velocity so that
the system do not have any chance to settle in a minimum of the potential.

More generally, as we already stated we believe that this ``dynamical''
approach to ML can be particularly fruitful when we want to consider
online learning problems, that is to say problems where the temporal
evolution of the parameters of the model at a certain stage of development 
depends explicitly on the data presented to the agent at the same time.
This means that it is particularly important to handle the case in which
the potential depends on time also trough a signal $u(t)$. Under this
assumption Eq.~\eqref{heavy} assumes the form
\[m \ddot w(t) +\friction\dot w(t) +\nabla U(w(t), u(t))=0.\]
This equation, in the limit $m\to 0$ yields
\[\dot w(t)=-\frac1 \friction\nabla U(w(t), u(t)),\]
that can be interpreted as the continuous
counterpart of a stochastic gradient descent method, when $u(t)$ is
interpreted as the realization of the random variable associated with
the data at the step $t$.
It is important to realize that whereas SGD is typically used in ML
assuming that the values of $u(t)$ are drawn from a training set
according to some probability distribution it is only when formulating the
problem using a signal $u(t)$ which has a temporal regularity (coherence)
that we can properly speak of online learning.

% \medskip
% \centerline{*\quad*\quad*}
% \medskip

% \noindent
% TO DO:
% \begin{itemize}
% \item Stress the importance of dissipation;
% \item Argue that, since we process a signal $u(t)$ we are interested in
% handling potentials that have a more general form $U(w(t), u(t))$
% \end{itemize}

% \medskip
% \centerline{*\quad*\quad*}
% \medskip

The paper is organized as follows: In Section~\ref{sec:2} we will show how
to reformulate least action principles in a more precise manner
following what has been done in~\cite{stefanelli}, Section~\ref{sec:3}
then shows how to extend some of the results of~\cite{stefanelli}
(namely the existence of the minimum for approximating problems) also in
the particularly interesting case where the potential explicitly
depends on time. Eventually Section~\ref{sec:4} closes the paper with some 
final considerations.

\def\+#1+{\vtop{\noindent\hsize 21pc \baselineskip9pt #1}}
\begin{table}[t]

\begin{tabular}{ccl}  
\toprule
{\bf Learning}  & {\bf Mechanics} & {\bf Remarks} \\
\midrule
$w$       & $\q$  &   \+Weights and neuronal outputs
                       are interpreted as generalized
                       coordinates+\\
\noalign{\smallskip}
$\dot w$& $\dot \q$ & \+ Weight variations and neuronal variations
                         are interpreted as generalized velocities.+\\
% \noalign{\medskip}
%   $A(w)$ & $S(\q)$& \+The cognitive action is the dual of the action in
%                       mechanics.+\\
\bottomrule
\end{tabular}
\caption{Links between learning theory and classical mechanics.}
\label{table}
\end{table}

\section{Lagrangian Mechanics}
\label{sec:2}
Following the approach proposed
in~\cite{stefanelli}, we will now discuss how it is possible to reformulate,
in a more precise manner,
the least action principle in classical mechanics. The following approach
can be directly applied, in the case of dissipative dynamics, to learning
processes simply through the identification of the generalized
coordinates of mechanics with the parameters of the learning
model (Table~\ref{table}).
In the remainder of the paper we will replace the variable $w$
which we used in the introduction to stress the connection with
the typical parameters (weights) used in ML with the generic coordinates
$\q$.

Usually (see~\cite{arnold} and~\cite{goldstein}) Hamilton's principle is formulated as
follows: Newton's laws of motion
\begin{equation}
\frac d{dt} (m\dot\q_i(t))+\nabla_i V(\q (t))=0,
\label{newton}
\end{equation}
coincide with extremals of the functional
\begin{equation}
\mathsf{S}(\q)
:=\int_0^T L\, dt,\quad \hbox{where}\quad L= \frac1 2 m|\dot\q|^2
-V(\q ),\label{action}\end{equation}
where $|\cdot|$ is the $n$-dimensional Euclidean norm.
This statement is usually also called {\it least action principle} even though
it is well known that the trajectory $\q (t)$ is not always a minimum
for the action. Another unsatisfactory
aspect of this principle is
the way in which the initial conditions are handled;
in newtonian mechanics Eq.~\eqref{newton}
is typically coupled with Cauchy initial conditions
\begin{equation}
\q (0)=\q^0,\qquad \dot\q (0)=\q^1,
\label{init-cond}
\end{equation}
that uniquely determine the motion of the system. On the other hand
Eq.~\eqref{newton} cannot be obtained from Hamilton's principle with
conditions~\eqref{init-cond}; usually the derivations make use of Dirichlet
boundary conditions (see~\cite{arnold}).

It has been shown (in \cite{stefanelli}) that Hamilton's principle can be
replaced by a {\it minimization} problem together with a limiting procedure.
In particular, let us consider the functionals
\begin{equation}\mathsf{W}_\varepsilon(\q):=\int_0^T e^{-t/\varepsilon}
\left(\frac{\varepsilon^2 m}2 |\ddot\q(t)|^2+V(\q(t))\right)\, dt,
\label{W_e}
\end{equation}
defined on the set $\dom\mathsf{W}_\varepsilon:=\{
\q\in H^2((0,T);\R^n)\mid \q(0)=\q^0, \dot\q(0)=\q^1\}$, where $V\in
\mathcal{C}^1(\R^n)$ and bounded from below and $m>0$.

The first property of this functional is that it admits a minimizer on
its domain; actually adding little bit of regularity on $V$ and choosing
$\varepsilon$ sufficiently small the minimizer turns out to be unique
(for a precise statement of this result see Lemma~4.1 of~\cite{stefanelli}).
Moreover the Euler-Lagrange equations for the minimizers 
of $\mathsf{W}_\varepsilon$ are (see Section~4 of~\cite{stefanelli})
\begin{align}
  &\varepsilon^2 m\q^{(4)}(t)-2\varepsilon m\q^{(3)}(t)
    +m\ddot \q(t)+\nabla V(\q(t))=0\quad t\in (0,T),\label{ELE-1}\\
  &\q(0)=\q^0,\quad \dot\q(0)=\q^1,\\
  &\ddot\q(T)=\q^{(3)}(T)=0.\label{end-cond}
\end{align}
Notice that from the stationarity condition of~\eqref{W_e} we get two
extra boundary conditions at time $t=T$ that seems to destroy causality
of the solution; one of the strengths of this approach however is that,
unlike Hamilton Principle, the boundary conditions~\eqref{end-cond} will
disappear in the limit $\varepsilon\to 0$ leaving the solution dependent only
on the initial state.

In the same limit ($\varepsilon\to 0$), we have that if $\q_\varepsilon$
solves~\eqref{ELE-1}--\eqref{end-cond}, then
(Theorem~4.2 of~\cite{stefanelli}) $\q_\varepsilon\to \q$ weakly in
$H^1((0,T);\R^n)$, where
$\q$ solves~\eqref{newton} with \eqref{init-cond}.
This last assertion makes clear that Hamilton principle can be reformulated
in terms of~\eqref{W_e} in the following way:
\begin{enumerate}
\item For each fixed $\varepsilon$ minimize $\mathsf{W}_\varepsilon$,
\item take the limit $\varepsilon\to 0$.
\end{enumerate}
Like Hamilton's principle this procedure is a variational approach
to classical mechanics, with respect to the principle of least
cognitive action however, as anticipated, it involves a true minimization
of the functional~\eqref{W_e} and it automatically reaches causality.

It is interesting to notice that if we omit step~2. in the procedure described 
above, stationarity conditions of~\eqref{W_e} would imply a 
dynamic based on differential equations of order higher than two (which has been 
actually considered in physics~\cite{suykens} and~\cite{nabulsi}). However the presence of 
the right boundary conditions~\eqref{end-cond} for each $\varepsilon>0$ 
would render the resulting laws non-causal.

To conclude this section we will discuss what can be considered yet  another
advantage of this approach by  showing how naturally it can handle
dissipative dynamics.

\medskip\noindent
{\bf Dissipative dynamics. } In the introduction we have briefly discussed how
dissipation is a fundamental feature for the formulation of learning
ad as a dynamical process; for this reason this point deserves a careful
discussion.

First of all notice that it is not possible to modify $L$ in Eq.~\eqref{action}
by choosing an appropriate $V$ or by adding additional derivative terms
in order to reproduce the following dissipative dynamics:
\begin{equation}
m\ddot\q+\friction\dot\q+\nabla V(\q)=0,
\label{dissipative-dyn}\end{equation}
with $\friction>0$. Nevertheless it has been shown (see~\cite{herrera}
and~\cite{tcs}) that it is possible to
include this kind of dynamic by the following modification of the action:
\[\mathsf{S}(\q)\to\overline{\mathsf{S}}(\q):=\int_0^Te^{\friction t/m}
\left(\frac1 2 m|\dot\q|^2 -V(\q )\right)\, dt.\]
This formulation changes the structure of the action functional making it more
similar to the $\mathsf{W}_\varepsilon$ functional. Still this
variational approach suffers of the same problems that has been discussed
previously in this section.

On the other hand in order to include dissipation in~\eqref{W_e}
it is sufficient
to modify the $\mathsf{W}_\varepsilon$ functional in the following way:
\[\mathsf{W}_\varepsilon(\q)\to\overline{\mathsf{W}}_\varepsilon(\q)
:=\int_0^T e^{-t/\varepsilon}
\left(\frac{\varepsilon^2 m}2 |\ddot\q(t)|^2+ \frac{\varepsilon\friction}2
|\dot\q(t)|^2+V(\q(t))\right)\, dt.\]
Then through the same minimization and limiting procedure described above
we recover Eq.~\eqref{dissipative-dyn} together with the correct
initial conditions~\eqref{init-cond}.

The modification
$\mathsf{W}_\varepsilon(\q)\to\overline{\mathsf{W}}_\varepsilon(\q)$ feels less
artificial than $\mathsf{S}(\q)\to\overline{\mathsf{S}}(\q)$ and  the term
added to $\mathsf{W}_\varepsilon$ seems a natural term to add. The reason
why the dissipative behaviour is recovered so easily by the variational approach
based on $\mathsf{W}_\varepsilon$ is that this principle is not invariant
by time reversal to begin with.

% Let us now choose $w(t)=\exp(-t/\varepsilon)$, $\mu=\varepsilon^2 m$,
% $\nu=\varepsilon \friction$ and  $\kappa=\friction_1=\friction_2=0$, with
% $m$, $\friction$ and $\varepsilon$ positive and real. Moreover, as it happens
% in mechanics, suppose that the potential depends on $t$ only though $\q$. 

\section{Generalization to time-dependent potential}
\label{sec:3}
The analysis presented in this section extends the result on the existence
of a minimizer to a family of functionals that include~\eqref{W_e}
where, in particular, we allow an explicit dependence on time
through the potential.

The following theory is relevant at least
for two distinct reason; first of all it is a first result  that goes in the
direction of extending the theory presented in~\cite{stefanelli}. In second
place it is interesting in its own (i.e. also if it is not coupled with a
limiting procedure) to ensure well-posedness of theories that relies on the
minimization of a functional of the form that we will consider. Recently
learning theories based on variational indexes considered in this section
has been used in Vision; in particular the proposed theory has been 
directly applied to the problem of feature extraction from a video signal $u(t)$ in 
an unsupervised manner with  the potential $U$ chosen to be  
 the mutual information between the 
visual data and a set of symbols (see~\cite{tnnls-vision}).

Let $T\in (0,\infty)$, $U\in C^0(\R^n\times\R^m)$ be bounded
from below such that $U(\cdot,0)\equiv 0$ and $\weight\in L^\infty(0,T)$
with $0<C_1\le \weight(t)\le C_2<+\infty$ for a.e. $t\in(0,T)$.
Let $u\colon [0,+\infty)
\to \R^m$ be an external input function that for the moment can be considered
a continuous function of time.
Consider the functional
\begin{equation}\Gamma(\q)=
  \int_0^T \weight(t)\Bigl(\frac{\mu}{2} |\ddot \q(t)|^2
  +\frac{\nu}{2}|\dot \q(t)|^2+\gamma\dot \q(t)\cdot\ddot \q(t)
  +\frac{\kappa }{2}|\q(t)|^2
  +U\bigl(\q(t),u(t)\bigr)\Bigr)\, dt, \label{functional}\end{equation}
where $\mu=\alpha+\gamma_2^2$, $\nu=\beta+\gamma_1^2$,
$\gamma=\gamma_1\gamma_2$, $\kappa >0$ are real numbers 
so that~\eqref{functional} can always be rewritten as
\[\Gamma(\q)=
  \int_0^T \weight(t)\Bigl(\frac{\alpha}{2} |\ddot \q(t)|^2
  +\frac{\beta}{2}|\dot \q(t)|^2+\frac{1}{2}|\gamma_1\dot \q(t)
  +\gamma_2\ddot \q(t)|^2
  +\frac{\kappa }{2}|\q(t)|^2
  +U\bigl(\q(t),u(t)\bigr)\Bigr)\, dt,\]
with $\alpha$, $\beta$, real and positive and $\q\in\dom(\Gamma):=
\{\,\q\in H^2((0,T); \R^n)
\mid \q(0)=\q^0, \quad \dot \q(0)=\q^1\,\}$, where $\q_0$, $\q_1\in \R^n$ are
given.

Suppose furthermore that we equip $\dom(\Gamma)$ with the following notion
of convergence:
\begin{equation}
\begin{aligned}
&\q_k\to \q\qquad \hbox{strongly in $H^1((0,T);\R^n)$};\\
&\ddot \q_k\rightharpoonup \ddot \q\qquad
\hbox{weakly in $L^2((0, T);\R^n)$}.\\
\end{aligned}
\label{convergence-notion}
\end{equation}
Then the following remark holds:

\begin{remark}
The set $\dom(\Gamma)$ is closed under the convergence in~\eqref{convergence-notion}, i.e., if
$\q_k\in\dom(\Gamma)$, $\q_k\to \q$ in $\dom(\Gamma)$, then $\q\in\dom(\Gamma)$.
\label{remark-A}
\end{remark}

Indeed, since $H^1(0,T)$ compactly embeds in $C([0,T])$ (see~\cite{brezis} 
pag. 213 Eq.~(6)) and
a weakly convergence sequence is strongly bounded (\cite{brezis} Prop.~3.5~(iii)), 
$\langle \q_k\rangle$ has a (not relabelled) subsequence
such that $\q_k\to \q$  and $\dot \q_k\to\dot \q$ uniformly in $[0,T]$,
therefore $\q(0)=\q^0$ and $\dot \q(0)=\q^1$.

% Notice that this notion of convergence is consistent with the definition
% of $K$, since the trace at $t=0$ of $q$ is preserved by convergence~(2).
% This follows from the following observation:

% \proclaim
% Observation A. By Sobolev immersion theorems $H^1(0\dts T)
% \mathrel{\mathop\subset^{\rm cpt}}  C([0\dts T])$ (see~[\Brezis],
% pag. 213 Eq.~(6)). This also mean that $H^2(0\dts T)
% \mathrel{\mathop\subset^{\rm cpt}} C^1([0\dts T])$, then if
% $\Vert q_n\Vert_{H^2(0\dts T)}\le C$ there exist a subsequence
% $q_{n_k}$ such that
% $$q_{n_k}\buildrel C^1([0\dts T])\over \ttto q\in C^1([0\dts T]).$$
% This means that if $q_n\to q$ in K, then $q_n(0)\to q(0)$ and
% $\dot q_n(0)\to \dot q(0)$.

We are now in the position to state the main result on the existence of
a minimum of the functional in~\eqref{functional}.

\begin{theorem}
The problem $\min\{\,\Gamma(\q)\mid \q\in\dom(\Gamma)\,\}$,
has a solution.\label{theorem}
\end{theorem}

\begin{proof}
We simply apply the direct method in the calculus of variations, namely
we have to show that $\Gamma$ is lower semicontinuous and coercive
with respect to the convergence in~\eqref{convergence-notion} and then we conclude in view of
Remark~\ref{remark-A}. \par{\it Lower semicontinuity.} The maps
$\q\in\dom(\Gamma)\mapsto \int \weight(t) |\q(t)|^2\, dt$
and $\q\in\dom(\Gamma)\mapsto \int \weight(t) |\dot \q(t)|^2\, dt$
are continuous, while $\q\in\dom(\Gamma)
\mapsto \int \weight(t) |\ddot \q(t)|^2\, dt$ is
lower semicontinuous (see~\cite{brezis} Prop.~3.5~(iii)); moreover
$\q\in\dom(\Gamma)\mapsto \int \weight(t) \dot \q(t)\cdot \ddot \q(t)\, dt$ is continuous because
of the strong-weak convergence of the scalar product in a  Hilbert space
(see~\cite{brezis}
Prop.~ 3.5~(iv)). Finally the map $\q\in\dom(\Gamma)
\mapsto \int \weight(t) U(\q(t), u(t))$ is
lower semicontinuous because of our assumptions on $U$ and as a
direct consequence of Fatou's Lemma. \par{\it Coercivity.}
Since $U$ is bounded from below and $T<\infty$ and in view of our assumptions
on $w$, $\alpha$, $\beta $, $\kappa $ it immediately follows that if  $\sup_{k\in\N}\Gamma(\q_k)<+\infty$, then there  exists a
constant $C>0$ such that $\Vert \q_k\Vert_{H^2}\le  C$ for
every $k\in\N$. Then from Theorem~3.16 in~\cite{brezis} it follows that
$\langle \q_k\rangle$ has a subsequence weakly converging in $H^2(0,T)$.
Moreover since $H^2(0,T)$ compactly embeds in $H^1(0,T)$ then
there is a subsequence that converges strongly in $H^1(0,T)$. This means
that indeed the sublevels of $\Gamma$ are compact with respect to the
convergence in Eq.~\eqref{convergence-notion}.
\end{proof}

\section{Conclusions}
\label{sec:4}
In this paper we presented an extension of the minimality result discovered
in~\cite{stefanelli} that entails the well-posedness of a class of learning
problems based on a Least Action Principle defined over the class of
functionals~\eqref{functional}. We prove that the existence of the minimum 
of $\Gamma$ (Theorem~\ref{theorem}) holds for a general weight 
function $\varpi$. Moreover, we argue that since learning requires
dissipation, the correspondent dynamics can be reproduced
from~\eqref{functional} by choosing $\varpi$ as an exponential function
of time, as discussed in Section~\ref{sec:2}. This paper provides
motivations to use the variational framework initially proposed in~\cite{tcs}, since it shows that, unlike the action of mechanics, the opportune
selections of the cognitive action leads to well-posed learning problems where
a global minimum can be discovered.

\begin{acknowledgement}
We thank Giovanni Bellettini for having brought to our attention 
the extended formulation of Newtonian mechanics and for insightful
discussions. 
\end{acknowledgement}
%
%\section*{Appendix}
%\addcontentsline{toc}{section}{Appendix}
%%
%%
%When placed at the end of a chapter or contribution (as opposed to at the end of the book), the numbering of tables, figures, and equations in the appendix section continues on from that in the main text. Hence please \textit{do not} use the \verb|appendix| command when writing an appendix at the end of your chapter or contribution. If there is only one the appendix is designated ``Appendix'', or ``Appendix 1'', or ``Appendix 2'', etc. if there is more than one.
%
%\begin{equation}
%a \times b = c
%\end{equation}

%%%%%%%%%%%%%%%%%%%%%%%% referenc.tex %%%%%%%%%%%%%%%%%%%%%%%%%%%%%%
% sample references
% %
% Use this file as a template for your own input.
%
%%%%%%%%%%%%%%%%%%%%%%%% Springer-Verlag %%%%%%%%%%%%%%%%%%%%%%%%%%
%
% BibTeX users please use
% \bibliographystyle{}
% \bibliography{}
%

\end{document}